\documentclass[sigconf]{acmart}

\usepackage[inline]{enumitem}

\usepackage{mathtools}
\usepackage[separate-uncertainty=true]{siunitx}

\DeclareMathOperator{\avg}{avg}

\newcommand{\mat}[1]{\boldsymbol{#1}}
\renewcommand{\vec}[1]{\boldsymbol{#1}}

\usepackage[capitalize,noabbrev]{cleveref}

\usepackage{xcolor}
\definecolor{rwth-blue}{RGB}{0,84,159}

\definecolor{rwth-magenta}{RGB}{227,0,102}
\definecolor{rwth-yellow}{RGB}{255,237,0}

\definecolor{rwth-petrol}{RGB}{0,97,101}
\definecolor{rwth-turquoise}{RGB}{0,152,161}
\definecolor{rwth-green}{RGB}{87,171,39}
\definecolor{rwth-maygreen}{RGB}{189,205,0}
\definecolor{rwth-orange}{RGB}{246,168,0}
\definecolor{rwth-red}{RGB}{204,7,30}
\definecolor{rwth-bordeaux}{RGB}{161,16,53}
\definecolor{rwth-violett}{RGB}{97,33,88}
\definecolor{rwth-purple}{RGB}{122,111,172}

\graphicspath{{figures/}}

\usepackage{tikz}  % not allowed
\usepackage{pgfplots}  % not allowed
\usepgfplotslibrary{dateplot, fillbetween, groupplots}
\usetikzlibrary{arrows.meta, backgrounds, calc, decorations.markings, intersections}

\tikzset{
    vertex/.style={circle, white, fill=rwth-blue, font={\scriptsize}, inner sep=0.5pt, minimum size=3pt},
	cell/.style={black, fill=gray, fill opacity=0.2},
}
\pgfplotsset{
	compat=1.18,
    compat/show suggested version=false,
    set layers,
	colormap={rwth}{
		color = (rwth-blue),
		color = (rwth-magenta),
		color = (rwth-orange),
		color = (rwth-petrol),
		color = (rwth-violett),
	},
    colormap name=rwth,
    every axis plot/.append style={line width=1.1pt},
	discard if not/.style 2 args={
		filter discard warning=false,
		x filter/.append code={
			\edef\tempa{\thisrow{#1}}
			\edef\tempb{#2}
			\ifx\tempa\tempb%
			\else
				
			\fi
		}
	}
}
\pgfplotscreateplotcyclelist{rwth}{
    [of colormap=rwth]
}
\pgfplotstableset{
    col sep=comma,
    discard row if not/.style 2 args={
        row predicate/.code={
            \def\pgfplotstable@loc@TMPd{\pgfplotstablegetelem{##1}{#1}\of}
            \expandafter\pgfplotstable@loc@TMPd\pgfplotstablename
            \edef\tempa{\pgfplotsretval}
            \edef\tempb{#2}
            \ifx\tempa\tempb
            \else
                \pgfplotstableuserowfalse
            \fi
        }
    }
}

\title{HLSAD: Hodge Laplacian-based Simplicial Anomaly Detection}

\author{Florian Frantzen}
\orcid{0000-0003-0187-3738}
\affiliation{%
    \institution{RWTH Aachen University}
    \department{Department of Computer Science}
    \city{Aachen}
    \country{Germany}
}
\email{florian.frantzen@cs.rwth-aachen.de}

\author{Michael T. Schaub}
\orcid{0000-0003-2426-6404}
\affiliation{%
    \institution{RWTH Aachen University}
    \department{Department of Computer Science}
    \city{Aachen}
    \country{Germany}
}
\email{schaub@cs.rwth-aachen.de}

\copyrightyear{2025}
\acmYear{2025}
\setcopyright{cc}
\setcctype{by}
\acmConference[KDD '25]{Proceedings of the 31st ACM SIGKDD Conference on Knowledge Discovery and Data Mining V.2}{August 3--7, 2025}{Toronto, ON, Canada}
\acmBooktitle{Proceedings of the 31st ACM SIGKDD Conference on Knowledge Discovery and Data Mining V.2 (KDD '25), August 3--7, 2025, Toronto, ON, Canada}
\acmDOI{10.1145/3711896.3736998}
\acmISBN{979-8-4007-1454-2/2025/08}


\begin{document}

\begin{abstract}
    In this paper, we propose HLSAD, a novel method for detecting anomalies in time-evolving simplicial complexes.
    While traditional graph anomaly detection techniques have been extensively studied, they often fail to capture changes in higher-order interactions that are crucial for identifying complex structural anomalies.
    These higher-order interactions can arise either directly from the underlying data itself or through graph lifting techniques.
    Our approach leverages the spectral properties of Hodge Laplacians of simplicial complexes to effectively model multi-way interactions among data points.
    By incorporating higher-dimensional simplicial structures into our method, our method enhances both detection accuracy and computational efficiency.
    Through comprehensive experiments on both synthetic and real-world datasets, we demonstrate that our approach outperforms existing graph methods in detecting both events and change points.
\end{abstract}

\begin{teaserfigure}
    \centering
    \includegraphics[width=\textwidth]{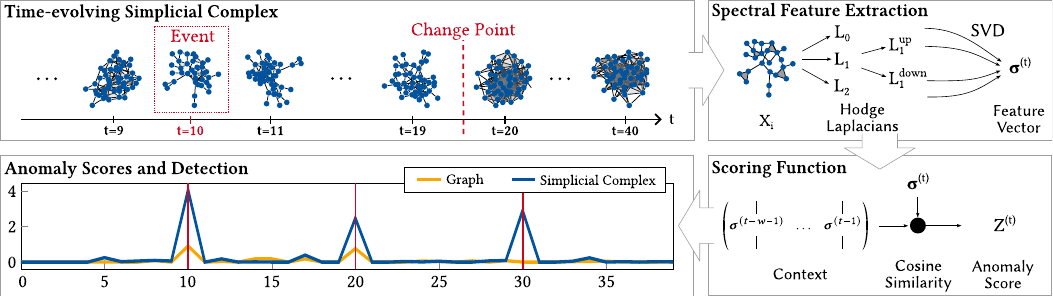}
    \caption{%
        \textbf{Illustrative process of anomaly detection on a toy-example simplicial complex sequence.}
        Starting from a time-evolving simplicial complex, we compute the Hodge Laplacians for each snapshot and extract spectral features for each.
        We employ a sliding window approach to identify anomalies by evaluating the deviation of current spectral features from established temporal patterns.
        In this illustrative example, HLSAD successfully detects all three artificially introduced anomalies, whereas traditional graph-based analysis fails to identify the final anomaly that manifests in higher-order interactions.
    }
    \label{figure:teaser}
    \Description{%
        Teaser figure illustrating HLSAD on a toy example.
        The figure shows a flowchart of the process:
        First, the input is a sequence of simplicial complexes, which are then processed to compute the Hodge Laplacians.
        Next, we extract spectral features from the Hodge Laplacians and collect them into context matrices for each timestep.
        These are processed into typical spectral features and an anomaly score is computed for each timestep by comparing the current spectral features to a characteristic feature vector.
        Finally, anomalies are detected based on the computed scores.
    }
\end{teaserfigure}

\maketitle

\newcommand\kddavailabilityurl{https://doi.org/10.5281/zenodo.15511585}

\ifdefempty{\kddavailabilityurl}{}{
    \begingroup\small\noindent\raggedright\textbf{KDD Availability Link:}\\
    The source code of this paper has been made publicly available at \url{\kddavailabilityurl} and at \url{https://git.rwth-aachen.de/netsci/hlsad}.
    \endgroup
}

\section{Introduction}

Anomaly detection within time-evolving networks has widespread applications across diverse domains, including network security for detecting intrusions and system faults~\citep{Ide:2004,Sun:2008,Ding:2012}, social network analysis for identifying collective behavioral changes induced by external factors~\citep{Peel:2015,Yu:2016}, and ecological monitoring for detecting ecosystem disturbances~\citep{Cheng:2008}.
Accordingly, many graph-based techniques have been developed for identifying events and change points in dynamic networks~\citep{Akoglu:2015,Ranshous:2015,Zhou:2024}.
Despite their utility, graph-based approaches are however limited by their focus on dyadic relationships, and fail to capture higher-order interactions that can be essential for identifying complex structural anomalies in the underlying system.
Such higher-order data can manifest itself through different mechanisms:
\begin{description}
    \item[Higher-Order Data]
          Many real-world datasets inherently contain multi-way interactions that extend beyond dyadic relationships.
          Traditional graph-based methods are fundamentally limited in their ability to capture and analyze these higher-order structures.
          For instance, during the COVID-19 pandemic, social distancing measures disproportionately affected large group gatherings compared to pairwise interactions, creating structural changes that are more readily detectable through higher-order methods~\citep{Pouw:2020}.

    \item[Lifting]
          Even when data only contains pairwise relations between entities, we can employ graph lifting techniques to construct a simplicial complex representation.
          A common approach identifies cliques in the graph with equivalent simplices.
          The effectiveness of lifting has been demonstrated in various contexts; notably, the Weisfeiler-Lehman isomorphism test exhibits strictly greater discriminative power when applied to clique complexes compared to their graph counterparts~\citep{Bodnar:2021a}.
\end{description}
Building upon the graph anomaly detection frameworks developed by \citet{Akoglu:2010} and \citet{Huang:2020}, we propose a novel method that leverages simplicial complexes to capture and analyze multi-way interactions in temporal data sequences.
It is important to distinguish our focus on temporal anomaly detection in time-evolving networks from the related but distinct problem of identifying anomalous substructures within static networks, as these two problems address fundamentally different analytical objectives despite their similar names.

The incorporation of higher-order interactions via simplices in our method provides significant advantages over traditional graph-based approaches:
First, it enhances detection accuracy through the explicit modeling of higher-order interactions that are inherently absent in graph representations.
These higher-order structures are particularly relevant in domains such as epidemic modeling, where large group gatherings may exhibit distinct response patterns to intervention measures compared to dyadic interactions.
Second, even if we only have pairwise interaction data, the application of graph lifting techniques results in improved anomaly detection capabilities.
Our experimental results indicate that our method achieves superior accuracy when applied to lifted graph datasets compared to conventional spectral approaches on graph skeletons and can even offer enhanced computational efficiency.

HLSAD operates through the following systematic process:
For each snapshot within a temporal sequence of simplicial complexes, we compute the corresponding Hodge Laplacians up to a user-specified order.
Subsequently, we extract the principal eigenvalues from both up and down Hodge Laplacians and concatenate these spectral features into a comprehensive feature vector for each temporal instance.
Employing a sliding window mechanism, we derive a characteristic feature vector by computing a weighted average across the window's temporal span.
When a new feature vector exhibits significant deviation from this established characteristic pattern, we designate the corresponding temporal instance as an anomaly.
We illustrate this systematic process in \cref{figure:teaser} and highlight a case where our higher-order approach successfully identifies an anomaly that is overlooked by traditional graph-based methods.

\paragraph{Contribution}

To the best of our knowledge, this paper presents the first methodology for anomaly detection in temporal sequences of simplicial complexes.
Our experimental results demonstrate that leveraging simplicial complex representations yields superior performance compared to traditional graph-based approaches in two key scenarios:
First, when analyzing datasets containing inherent higher-order interactions, and second, when applying graph lifting techniques to enrich the topological structure.

\paragraph{Outline}

The remainder of this paper is organized as follows:
In \cref{section:related-work}, we review the existing literature on graph anomaly detection methods.
In \cref{section:preliminaries}, we introduce the mathematical framework and notation used throughout this work.
\Cref{section:problem-statement} formally defines the anomaly detection problem on simplicial complexes that we focus on.
In \cref{section:method}, we present our spectral approach for detecting anomalies using Hodge Laplacians.
\Cref{section:experiments} evaluates our method on synthetic and real-world datasets, demonstrating its effectiveness for event and change point detection.
Finally, \cref{section:conclusion} summarizes our findings and discusses future research directions.

\section{Related Work}%
\label{section:related-work}

Following the taxonomy established by \citet{Ranshous:2015}, existing graph anomaly detection algorithms can be categorized into five principal categories:
\begin{enumerate*}
    \item community-based,
    \item compression-based,
    \item decomposition-based,
    \item similarity/distance-based, and
    \item probabilistic model-based approaches.
\end{enumerate*}
Many methods can be subsumed under the following common framework:
Initially, the algorithm extracts a low-dimensional feature representation from each temporal snapshot.
Subsequently, these feature vectors are analyzed to establish baseline dissimilarity metrics that characterize normal temporal fluctuations of the data in the absence of an anomaly.
When a snapshot exhibits dissimilarity exceeding an established threshold, it is identified as an anomaly.
Events and change points can be detected by whether the shift in dissimilarity is localized or persists over the following snapshots.

Further, anomaly detection algorithms can be characterized by several essential properties, with existing methods exhibiting varying combinations of these attributes~\citep{Huang:2020,Wang:2017}:
First, the approach should maintain generality without imposing assumptions about the underlying data distribution, such as conformity to a specific graph model.
Second, algorithms should be applicable to both event and change point detection scenarios.
Third, computational scalability is crucial, particularly for large-scale networks with dynamic node populations.
Within this context, we can also consider offline and online-style algorithms, where the former analyzes the entire time-series at once, while the latter flags anomalies in real-time as new data arrives.
Fourth, node permutation invariance is essential to accommodate datasets where consistent node ordering across temporal snapshots cannot be guaranteed.
Fifth, the method should provide quantitative confidence measures for anomaly predictions, enhancing interpretability and decision-making capabilities.

Model-based detection algorithms are a prominent approach in literature, wherein probabilistic graph models are fitted to observed data and anomalies are identified through parameter changes.
These approaches offer significant advantages in terms of interpretability, as detected parameter shifts directly correspond to identifiable large-scale structural changes in the network.
\citet{Peel:2015} developed an online learning algorithm for change point detection in dynamic networks, learning probabilistic distributions over graphs and detecting when these distributions shift due to external events.
Their method utilizes a generalized hierarchical random graph model for temporal network representation, though the framework remains model-agnostic.
Notably, they identified a list of different types of change points, including community formation and merge events.
In a similar direction, \citet{Wang:2017} proposed EdgeMonitoring, which derives snapshot features through joint edge probability estimation.
Their method quantifies temporal dissimilarity using Kolmogorov-Smirnov statistics and Kullback-Leibler divergence measures.
\Citet{Gahrooei:2018} developed a framework that fits generalized linear models to sequential network snapshots, employing extended Kalman filtering for continuous parameter estimation.
They identify structural changes through significant deviations in the estimated model parameters.

\Citet{Ide:2004} (Activity Vector), \citet{Akoglu:2010} and \citet{Huang:2020} (LAD) proposed different spectral-based methods for anomaly detection in dynamic graphs, which all rely on identifying changes in the graph spectrum as proxy for structural changes.
These methods have been shown to be effective in detecting both event and change points in dynamic networks.
In particular, LAD was later extended to attributed \citep{Huang:2023} and multi-view graphs \citep{Huang:2024}.

Complementing these approaches, \citet{Koutra:2012} developed TENSORSPLAT, a decomposition-based method that leverages tensor factorization to identify structural changes in dynamic networks.
\Citet{Koutra:2016} introduced \textsc{DeltaCon}, which utilizes direct similarity-based comparisons between consecutive graph snapshots rather than utilizing intermediate feature representations.
While this method offers an alternative perspective on structural change detection, it imposes significant constraints: the framework requires both a static node set and consistent node identification across temporal snapshots.

With the increased interest in deep learning approaches in recent years, several methods have emerged that leverage neural network architectures for anomaly detection in dynamic graphs.
Notably, \citet{Gong:2023} conceptualized change point detection as a temporal prediction framework, employing a latent evolution model to forecast subsequent graph snapshots.
Their approach identifies anomalies when significant discrepancies arise between predicted and observed network states.
\Citet{Sulem:2024} proposed a method based on siamese graph neural networks to learn a similarity function between graph snapshots and detect change points based on deviations from this learned similarity.
To the best of our knowledge, however, again no deep learning approach explores higher-order data in a simplicial complex setting.

For an in-depth review of existing anomaly detection methods on dynamic graphs, readers are directed to the recent surveys by \citet{Zhou:2024} and by \citet{Ho:2025}.
Additional foundational perspectives can be found in the older surveys by \citet{Akoglu:2015} and \citet{Ranshous:2015}.

\section{Preliminaries}%
\label{section:preliminaries}

\paragraph{Notation}

We denote vectors with boldface lowercase letters $\vec{v}$ and matrices with boldface capital letters $\mat{F}$ throughout this work.
The temporal index of a variable at time step $t$ is indicated by the superscript notation $(\cdot)^{(t)}$.

\paragraph{Simplicial Complexes}

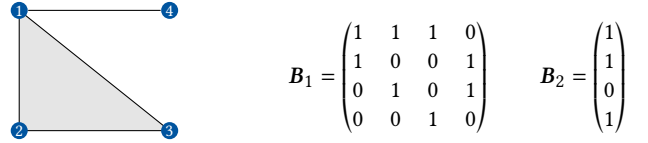
\begin{figure}[t]
    \begin{minipage}{0.35 \linewidth}
        \begin{tikzpicture}
            \node[vertex] (1) at (0, 1.6) {$1$};
            \node[vertex] (2) at (0, 0) {$2$};
            \node[vertex] (3) at (2, 0) {$3$};
            \node[vertex] (4) at (2, 1.6) {$4$};

            \begin{pgfonlayer}{axis ticks}
                \draw[cell] (1.center) -- (2.center) -- (3.center) -- cycle;
                \draw (1) -- (4);
            \end{pgfonlayer}
        \end{tikzpicture}
    \end{minipage}
    \hfill
    \begin{minipage}{0.60 \linewidth}
        \begin{align*}
            \mat{B}_1 & = \begin{pmatrix}
                              1 & 1 & 1 & 0 \\
                              1 & 0 & 0 & 1 \\
                              0 & 1 & 0 & 1 \\
                              0 & 0 & 1 & 0
                          \end{pmatrix} &  &  &
            \mat{B}_2 & = \begin{pmatrix}
                              1 \\
                              1 \\
                              0 \\
                              1
                          \end{pmatrix}
        \end{align*}
    \end{minipage}
    \caption{%
        \textbf{Illustration of a simplicial complex with associated boundary matrices.}
        The simplicial complex contains a $2$-simplex as shaded area.
        The rows and columns of the boundary matrices are ordered lexicographically.
    }
    \label{figure:simplicial-complex}
    \Description{%
        A simplicial complex with four vertices, placed into a triangle and one edge attached to one of the triangle's vertices.
        The triangle is a $2$-simplex, while the edge is a $1$-simplex.
        The unsigned boundary matrices $B_1$ and $B_2$ are shown, where $B_1$ is the vertex-to-edge incidence matrix and $B_2$ is the edge-to-triangle incidence matrix.
    }
\end{figure}

Given a set of points $\mathcal{V}$, a $k$-simplex $\mathcal{S}^k$ is a subset of $k+1$ points.
An (abstract) simplicial complex (SC) $\mathcal{X}$ comprises a collection of simplices that satisfies the following simplicial closure property: for any $k$-simplex $\mathcal{S}^k \in \mathcal{X}$, all of its proper subsets $\mathcal{S}^{k'} \subsetneq \mathcal{S}^k$ with $0 < k' < k$ are also elements of $\mathcal{X}$.
We denote the set of $k$-simplices in $\mathcal{X}$ by $\mathcal{X}_k$.
The rank $k_\text{max}$ of a simplicial complex is defined as the maximum dimension among all its simplices, i.e., $k_\text{max} = \max \{ \vert \mathcal{S}^k \vert \mid \mathcal{S}^k \in \mathcal{X} \}$.
Following standard terminology for graphs, we refer to $0$-simplices as vertices, $1$-simplices as edges, and $2$-simplices as triangles.

The algebraic structure of a simplicial complex can be encoded through boundary matrices $\mat{B}_k$, which formalize the incidence relationships between simplices of adjacent dimensions~\citep{Bredon:1993,Hatcher:2002}.
Specifically, the rows of $\mat{B}_k$ correspond to $(k-1)$-simplices and columns correspond to $k$-simplices.
The entry $(i,j)$ takes value $1$ if the $i$-th $(k-1)$-simplex is incident to the $j$-th $k$-simplex, and $0$ otherwise.
In particular, $\mat{B}_1$ represents the (unsigned) vertex-to-edge incidence matrix, $\mat{B}_2$ captures the edge-to-triangle incidence relations, and so on.
For notational convenience, we define $\mat{B}_0 = \mat{B}_{k_\text{max}+1} = \mat{0}$.
We illustrate these constructs in \cref{figure:simplicial-complex}.

Utilizing these boundary matrices, we define the $k$-th combinatorial Hodge Laplacian $\mat{L}_k$ of the simplicial complex $\mathcal{X}$ as follows:
\begin{align}
    \mat{L}_k & = \mat{B}_k^\top \mat{B}_k + \mat{B}_{k+1} \mat{B}_{k+1}^\top.
\end{align}
Notably, when $k=0$, we recover the classical graph Laplacian $\mat{L}_0 = \mat{B}_{k+1} \mat{B}_{k+1}^\top$.
To facilitate our analysis, we decompose the Hodge Laplacian into its constituent components: the down-Laplacian $\mat{L}_k^\text{down} = \mat{B}_{k}^\top \mat{B}_{k}$ and the up-Laplacian $\mat{L}_k^\text{up} = \mat{B}_{k+1} \mat{B}_{k+1}^\top$, which capture distinct aspects of the complex's structure.

In practical applications, it is often advantageous to assign non-negative weights to simplices to quantify their relative importance in the complex.
Let $w_k \colon \mathcal{X}_k \to \mathbb{R}_{\geq 0}$ denote the weight function for $k$-simplices and $\mat{W}_k$ the corresponding diagonal weight matrix.
The boundary operators can be generalized to incorporate these weights by defining the weighted boundary matrix $\tilde{\mat{B}}_k = \mat{W}_{k-1}^{-1} \mat{B}_k \mat{W}_k$~\citep{Guglielmi:2023}.
Consequently, the weighted $k$-th Hodge Laplacian takes the form:
\begin{align}
    \tilde{\mat{L}}_k & = \mat{W}_k \mat{B}_k^\top \mat{W}_{k-1}^{-2} \mat{B}_k \mat{W}_k + \mat{W}_k^{-1} \mat{B}_{k+1} \mat{W}_{k+1}^2 \mat{B}_{k+1} \mat{W}_k^{-1}.
\end{align}

\paragraph{Synthetic Model and Continuity Rate}

Consider a temporal sequence of graphs and an underlying generative model that governs the evolution of the graph structure over time.
The temporal evolution between consecutive graph snapshots is regulated by the continuity rate $\alpha^{(t)}$~\citep{Wang:2017}, which determines the balance between preserved and resampled dyadic relationships.
Specifically, for each dyad $e$, the probability $1 - \alpha^{(t)}$ governs whether its connection status persists from time step $t-1$, while with probability $\alpha^{(t)}$ its status is resampled according to the current generation model.
Thus, the structural similarity between adjacent temporal snapshots exhibits an inverse relationship with $\alpha^{(t)}$, where lower values yield higher topological preservation.

\begin{figure}
    \resizebox{\linewidth}{!}{
        \begin{tikzpicture}[
        label-node/.style={anchor=west, align=left},
        variable-node/.style={minimum size=0.75cm, font=\small},
        observed/.style={variable-node, circle, draw},
        latent/.style={observed, gray}
    ]

    \begin{scope}
        \node[label-node, font=\Large] at (0, 2.5) {\textbf{A: Simplicial Data}};

        \node[label-node, gray] (latent-label) at (0,1.5) {Latent Generative \\ SC Model};
        \node[label-node] (sc-label) at (0,0) {Observed \\ SCs};

        % Model variables
        \node[latent] (M1) at (3.5,1.5) {$M^{(1)}$};
        \node[latent] (M2) at (5.5,1.5) {$M^{(1)}$};
        \node[latent] (M3) at (7.5,1.5) {$M^{(3)}$};
        \node[gray, variable-node] (Mt) at (9,1.5) {$\cdots$};

        \draw[->, gray] (M1) -- (M2);
        \draw[->, gray] (M3) -- (Mt);

        % SC variables
        \node[observed] (X1) at (3.5,0) {$\mathcal{X}^{(1)}$};
        \node[observed] (X2) at (5.5,0) {$\mathcal{X}^{(2)}$};
        \node[observed] (X3) at (7.5,0) {$\mathcal{X}^{(3)}$};
        \node[variable-node] (Xt) at (9,0) {$\cdots$};

        % Time
        \node at (3.5, -0.7) {$t = 1$};
        \node at (5.5, -0.7) {$t = 2$};
        \node at (7.5, -0.7) {$t = 3$};

        % Vertical arrows
        \draw[->, gray] (M1) -- (X1);
        \draw[->, gray] (M2) -- node[midway, right, black] {$\alpha^{(2)}$} (X2);
        \draw[->, gray] (M3) -- node[midway, right, black] {$\alpha^{(3)}$} (X3);

        % Horizontal arrows
        \draw[->] (X1) -- node[midway, above] {$1 - \alpha^{(2)}$} (X2);
        \draw[->] (X2) -- node[midway, above] {$1 - \alpha^{(3)}$} (X3);
        \draw[->] (X3) -- (Xt);

        % Dashed line between latent and observed
        \draw[dashed, gray, opacity=0.5] ($(latent-label.west)!0.5!(sc-label.west)$) -- ($(Mt.east)!0.5!(Xt.east)$);

        % Anomaly
        \draw[red] ($(M2.north)!0.5!(M3.north)$) -- ($(M2.south)!0.5!(M3.south)$);
        \node[red] at ($(M2.north)!0.5!(M3.north) + (0, 0.25)$) {Anomaly};
    \end{scope}

    \begin{scope}[yshift=-5.5cm]
        \node[label-node, font=\Large] at (0, 4.0) {\textbf{B: Graph Lifting}};

        \node[label-node, gray] (latent-label) at (0,3.0) {Latent Generative \\ Graph Model};
        \node[label-node, gray] (graph-label) at (0,1.5) {Intermediate \\ Graphs};
        \node[label-node] (sc-label) at (0,0) {Lifted \\ SCs};

        % Model variables
        \node[latent] (M1) at (3.5,3.0) {$M^{(1)}$};
        \node[latent] (M2) at (5.5,3.0) {$M^{(1)}$};
        \node[latent] (M3) at (7.5,3.0) {$M^{(3)}$};
        \node[gray, variable-node] (Mt) at (9,3.0) {$\cdots$};

        \draw[->, gray] (M1) -- (M2);
        \draw[->, gray] (M3) -- (Mt);

        % Graph variables
        \node[latent] (G1) at (3.5,1.5) {$G^{(1)}$};
        \node[latent] (G2) at (5.5,1.5) {$G^{(2)}$};
        \node[latent] (G3) at (7.5,1.5) {$G^{(3)}$};
        \node[variable-node] (Gt) at (9,1.5) {$\cdots$};

        % SC variables
        \node[observed] (X1) at (3.5,0) {$\mathcal{X}^{(1)}$};
        \node[observed] (X2) at (5.5,0) {$\mathcal{X}^{(2)}$};
        \node[observed] (X3) at (7.5,0) {$\mathcal{X}^{(3)}$};
        \node[variable-node] (Xt) at (9,0) {$\cdots$};

        % Time
        \node at (3.5, -0.7) {$t = 1$};
        \node at (5.5, -0.7) {$t = 2$};
        \node at (7.5, -0.7) {$t = 3$};

        % Vertical arrows
        \draw[->, gray] (M1) -- (G1);
        \draw[->, gray] (M2) -- node[midway, right, black] {$\alpha^{(2)}$} (G2);
        \draw[->, gray] (M3) -- node[midway, right, black] {$\alpha^{(3)}$} (G3);

        % Vertical arrows
        \draw[->, gray] (G1) -- (X1);
        \draw[->, gray] (G2) -- (X2);
        \draw[->, gray] (G3) -- (X3);

        % Horizontal arrows
        \draw[->, gray] (G1) -- node[midway, above, black] {$1 - \alpha^{(2)}$} (G2);
        \draw[->, gray] (G2) -- node[midway, above, black] {$1 - \alpha^{(3)}$} (G3);
        \draw[->, gray] (G3) -- (Gt);

        % Dashed line between latent and observed
        \draw[dashed, gray, opacity=0.5] ($(graph-label.west)!0.5!(sc-label.west)$) -- ($(Gt.east)!0.5!(Xt.east)$);

        % Anomaly
        \draw[red] ($(M2.north)!0.5!(M3.north)$) -- ($(M2.south)!0.5!(M3.south)$);
        \node[red] at ($(M2.north)!0.5!(M3.north) + (0, 0.25)$) {Anomaly};
    \end{scope}
\end{tikzpicture}
    }
    \caption{
        \textbf{Illustration of the continuity rate $\alpha^{(t)}$ governing the temporal evolution of a simplicial complex in a generative model setting}.
        The latent variables $M^{(t)}$ denote the underlying generative model, while $\mathcal{X}^{(t)}$ represents the observed simplicial complex snapshots.
        In (A), we show the case of underlying higher-order data, while (B) illustrates the scenario of graph lifting.
        The continuity rate $\alpha^{(t)}$ determines the proportion of simplices or edges that persist versus those resampled between consecutive temporal instances.
        A change in the model parameters at time step $3$ illustrates an anomalous event that the detection algorithm aims to identify.
    }%
    \label{figure:continuity-rate}
    \Description{%
        Shows the information flow in a generative model setting for our anomaly detection.
        The top shows the case of higher-order data, in which a latent generative simplicial complex model generates simplicial complexes.
        The model changes at time step $3$, which is an anomaly that the detection algorithm should identify.
        The observed simplicial complex snapshots depend on the current generative model and the previous snapshot, as governed by the continuity rate.
        The bottom shows the case of graph lifting, where the latent model generates graph skeletons only and we apply a graph lifting technique to obtain higher-order simplices.
        In this case, the continuity rate acts on the intermediate graph skeletons, not the final simplicial complexes.
    }
\end{figure}
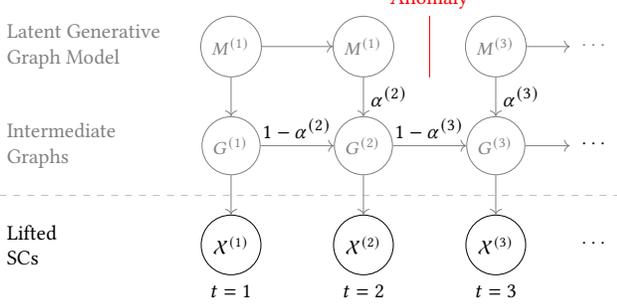

The extension of continuity rates to simplicial complexes presents significant challenges due to the inherent dependencies between simplices of different ranks:
The removal of an edge that forms part of a triangle's boundary necessitates the removal of the triangle, regardless of the continuity rate acting upon it.
Conversely, when a resampled edge completes a triangle's boundary that was previously incomplete, it creates the potential for a new triangle.
We address these dependencies through two distinct approaches, determined by the origin of the higher-order interactions:

For graph lifting scenarios, we apply the continuity rate exclusively to the graph skeleton.
Higher-order simplices are derived through clique lifting, independent of the continuity rate.

For data-informed higher-order simplices, we restrict the application of the continuity rate to simplex candidates that maintain validity across consecutive snapshots.
Invalid simplices are systematically removed, while new simplex candidates are always sampled from the generation model.
This approach facilitates the emergence of new higher-order simplices in subsequent snapshots, a practical consideration for evolving complex structures.

\Cref{figure:continuity-rate} illustrates the continuity rate $\alpha^{(t)}$ governing the temporal evolution of a sequence of simplicial complexes in a generative model setting.

\section{Problem Statement}%
\label{section:problem-statement}

Let $\mathcal{X}$ denote a dynamic simplicial complex represented as a sequence of snapshots $\{ \mathcal{X}^{(t)} \}_{t=1}^T$.
Each simplex exists at specific timestamps $t$ and may dynamically appear or disappear throughout the time series.
Furthermore, we do not require identification of simplices across temporal instances, permitting arbitrary vertex and simplex orderings within each snapshot.
Unlike some graph-based methods that require a fixed number of vertices~\citep{Koutra:2012,Wang:2017}, our formulation does not impose such constraints.
Specifically, we allow for a varying number of nodes across timestamps and do not require knowledge of the total number of vertices in the system.

Based on this formulation, our objective is to identify anomalous simplicial complexes $\mathcal{X}^{(t)}$ within the time series $\mathcal{X}$.
Specifically, given an anomaly scoring function $f \colon \mathcal{X} \to \mathbb{R}$, we aim to detect time steps $t$ exhibiting anomalous behavior.
Following the taxonomy proposed by \citet{Ranshous:2015}, we differentiate between two distinct types of anomalies:
\begin{itemize}
    \item
          \emph{Event points} represent temporal instances where a snapshot exhibits significant deviation from the typical behavior of the system.
          These deviations may arise from various sources including sensor errors, data corruption, or external perturbations.
          More formally, we characterize an event point $t$ by the condition $| f(\mathcal{X}^{(t)}) - f(\mathcal{X}(W)) | > \epsilon$, where $\mathcal{X}(W)$ denotes the short-term behavior within a recent context window $W$ and $\epsilon$ represents a significance threshold.

    \item
          \emph{Change points} indicate temporal locations where fundamental shifts occur in the underlying data distribution.
          Such transitions may result from external events altering the environment or modifications in the intrinsic data generation process.
          Formally, a change point $t$ satisfies $| f(\mathcal{X}^{(t)}) - f(\mathcal{X}_\text{norm}) | > \delta$, where $\mathcal{X}_\text{norm}$ captures the normal behavior in the global context and $\delta$ denotes a threshold parameter.
\end{itemize}
The anomaly scoring function should provide clear discrimination between anomalous and normal points while maintaining monotonicity with respect to the degree of anomalousness.
We note that in the literature, event points are sometimes referred to as anomaly points, and a unified terminology for both types of anomalies is not always consistently employed.

\section{Method}%
\label{section:method}

In this section, we present our method HLSAD for detecting event and change points in temporal sequences of simplicial complexes.
This essentially implements the anomaly scoring function $f$ introduced in the previous section, which assigns an anomaly score to each snapshot by analyzing its deviation from both short-term and long-term temporal contexts, defined by window sizes $w_s$ and $w_\ell$ respectively.

On a high level, HLSAD consists of the following steps:
First, given an input time series $\{ X^{(t)} \}_{t=1}^T$ of simplicial complexes, we compute the combinatorial Hodge Laplacians $\mat{L}_k$ up to a user-specified maximum rank $K$ for each temporal snapshot.
These Laplacians capture the topological information at different dimensional levels of the simplicial complex.
Second, we extract spectral features by computing the $\ell$ largest singular values from each Hodge Laplacian, concatenating them into a feature vector.
This step effectively compresses the high-dimensional structural information into a lower-dimensional representation while preserving the most significant topological characteristics.
Finally, we employ a sliding window approach to detect anomalies by comparing the current feature vector against both short-term and long-term temporal contexts.
Significant deviations from these temporal contexts are flagged as potential anomalies.

\subsection{Spectrum}

HLSAD leverages spectral features obtained through singular value decomposition (SVD) of the Hodge Laplacians for each temporal snapshot, extending established graph-based techniques to the simplicial domain.
The choice of singular values as discriminative features is motivated by several theoretical considerations:

First, the Hodge Laplacian spectrum captures fundamental structural and topological properties of the simplicial complex: for instance, the dimension of the kernel space of the $k$-th Hodge Laplacian equals the $k$-th Betti number of the complex~\citep{Eckmann:1944}.
This provides information about its topological features such as connected components, cycles, and holes.

Second, by considering both up and down Laplacians at each dimension, we capture distinct aspects of the simplicial structure.
The up-Laplacian $\mat{L}_k^\text{up}$ encodes how $k$-simplices combine to form $(k+1)$-simplices, while the down-Laplacian $\mat{L}_k^\text{down}$ describes how $k$-simplices decompose into $(k-1)$-simplices.
This bidirectional analysis provides a comprehensive view of the complex's structure that would be impossible to achieve through graph-based methods.

Third, truncated SVD provides optimal low-rank approximations of matrices with respect to both the Frobenius and $2$-norm~\citep{Eckart:1936}.
The $(k+1)$-th singular value $\sigma_{k+1}$ quantifies the reconstruction error of the best rank-$k$ approximation in the $2$-norm.
Thus, the ordered singular spectrum $\sigma_1, \sigma_2, \dots, \sigma_r$ encodes the information loss that would occur at different levels of approximation.
Significant changes in these values indicate structural modifications that affect the complex's topology.

Fourth, our method maintains permutation invariance with respect to the ordering of simplices.
This property is crucial for practical applications where consistent ordering cannot be guaranteed across temporal snapshots.
Row or column permutations of the Hodge Laplacian matrices do not affect their singular values, enabling our method to handle arbitrary simplex orderings.

Finally, the computational efficiency of sparse SVD algorithms makes our approach practical for large-scale datasets.
Real-world simplicial complexes often exhibit sparsity, particularly at higher dimensions where the number of simplices typically decreases.
By computing only the top $k$ singular values through truncated SVD, we can achieve significant computational savings compared to full decomposition while maintaining high detection accuracy.

The selection of spectral features from Hodge Laplacians presents multiple strategic approaches, each capturing distinct topological characteristics of the simplicial complex while offering different computational trade-offs.
These methodological choices significantly influence both the structural information preserved and the algorithmic efficiency of the detection process and can be tuned on a per-dataset basis.
We discuss this decision matrix further in the next section.
For simplicity, here we use the first $K$ Hodge Laplacians and extract the first $\ell$ singular values from each, resulting in a feature vector of length $K \cdot \ell$ for each temporal snapshot.
If a simplicial complex has a rank lower than $K$, or if the number of singular values is less than $\ell$, we pad the feature vector with zeros.

\subsection{Selection of Singular Values}%
\label{section:selection-of-singular-values}

The selection of singular values exhibits multiple dimensions of choice:
First, we can compute singular values for different ranks $k$ of Hodge Laplacians, enabling analysis of distinct structural properties within the simplicial complex.
In our implementation, we consider the first $K$ Hodge Laplacians, which is motivated by two key observations:
For one, dyadic interactions serve as fundamental indicators for anomaly detection, and second, these lower-order interactions inherently influence higher-order connections through the simplicial closure property.
From a practical perspective, simplicial complexes typically exhibit increasing sparsity at higher dimensions, allowing for the selection of a relatively small value of $K$ without significant loss of structural information.

Second, the selection of $\ell$ singular values per Hodge Laplacian determines the dimensionality of the spectral representation.
This parameter establishes a fundamental trade-off between computational efficiency and structural information retention:
A larger value of $\ell$ enables us to capture more nuanced topological features at the cost of increased computational complexity of the anomaly detection algorithm.
Retaining only a relatively small number of singular values usually sufficiently characterizes the dominant structural properties while maintaining computational tractability.
While one could theoretically optimize this parameter independently for each order, our code and experiments maintain a uniform selection across ranks to reduce the parameter space complexity.

Third, in contrast to the graph Laplacian, the Hodge Laplacian is composed of two distinct components: the down-Laplacian $\mat{L}_k^\text{down}$ and the up-Laplacian $\mat{L}_k^\text{up}$ (with the graph Laplacian consisting solely of the latter component).
This decomposition introduces an additional dimension for spectral feature optimization, enabling selective focus on the spectral properties of either component individually or a combination of both.
This way, depending on the dataset, one can fine-tune the algorithm to emphasize different structural aspects of the simplicial complex.

\subsection{Context Matrix and Typical Behavior}

Following established practices in anomaly detection literature, we evaluate anomalies relative to a temporal context window of size $w$ containing previous observations.
To that end, we maintain a context matrix $\mat{C}^{(t)}$ comprising the $\ell_2$-normalized spectra from the last $w$ snapshots:
\begin{align}
    \mat{C}^{(t)} & = \begin{pmatrix}
                          \mid                   & \mid                 &        & \mid                 \\
                          \vec{\sigma}^{(t-w-1)} & \vec{\sigma}^{(t-w)} & \ldots & \vec{\sigma}^{(t-1)} \\
                          \mid                   & \mid                 &        & \mid
                      \end{pmatrix} \in \mathbb{R}^{n \times w}.
\end{align}
This context matrix enables derivation of a “typical” spectrum $\tilde{\vec{\sigma}}^{(t)}$ over the preceding $w$ snapshots, which serves as a baseline for comparison with the current snapshot $\vec{\sigma}^{(t)}$.
\citet{Akoglu:2010} investigated two methodologies for obtaining such a typical feature vector:
First, they used
\begin{align}
    \tilde{\vec{\sigma}}^{(t)} = \frac{1}{w} \sum_{i=1}^w \mat{C}_{:,i} = \frac{1}{w} \sum_{i=1}^{w} \vec{\sigma}^{(t-i-1)}
\end{align}
as the arithmetic average of the last $w$ feature vectors.
Second, they computed the left singular vector of the context matrix $\mat{C}^{(t)}$ using SVD decomposition, which can be interpreted as a weighted average of the preceding feature vectors.
Both their empirical analysis and our experiments demonstrated superior performance of the SVD-based approach, leading us to adopt the SVD-based averaging.

\subsection{Scoring Function}

Let $\tilde{\vec{\sigma}}_s^{(t)}$ and $\tilde{\vec{\sigma}}_\ell^{(t)}$ denote the characteristic spectral features derived according to the previous section within the short-term and long-term context windows $w_s$ and $w_\ell$, respectively.
Our goal now is to quantify the dissimilarity between the current spectral features $\vec{\sigma}^{(t)}$ and these characteristic behaviors.

To quantify the structural deviation, we employ the angular distance between the current feature vectors and the established typical spectra:
\begin{align}
    Z_s^{(t)}    & = 1 - \left(\vec{\sigma}_s^{(t)}\right)^\top \tilde{\vec{\sigma}}_s^{(t)}        \\
    Z_\ell^{(t)} & = 1 - \left(\vec{\sigma}_\ell^{(t)}\right)^\top \tilde{\vec{\sigma}}_\ell^{(t)}.
\end{align}
This normalized dissimilarity measure yields values in the interval $[0,1]$, where $Z=0$ corresponds to perfect structural alignment between current and characteristic features, while $Z=1$ indicates maximal topological deviation manifested through orthogonality of the feature vectors.
We define the final anomaly score
\begin{align}
    Z^{(t)} & = \max \{ Z_s^{(t)}, Z_\ell^{(t)} \}
\end{align}
as the maximum deviation across both short-term and long-term contexts.

Having established the anomaly scoring function, we can employ multiple strategies to identify events and change points:
The first approach establishes a fixed threshold $\tau$ and classifies temporal instances $t$ where $Z^{(t)} > \tau$ as anomalies.
This threshold can be calibrated using an initial training period, during which we assume normal system behavior, by analyzing the distribution of $Z^{(t)}$ scores for a fixed “initialization window” at the beginning of the time series.
Alternatively, we can adopt a rank-based approach that identifies the $k$ snapshots with the highest anomaly scores.
This latter method proves particularly effective when focusing on the most significant structural deviations, as it automatically adapts to the scale of the anomaly scores without requiring explicit threshold calibration.

Having defined our anomaly scoring function $Z^{(t)}$, we can differentiate between event and change points by analyzing the persistence of high scores across consecutive time steps:
Event points typically manifest as isolated spikes in the anomaly score, where $Z^{(t)}$ exhibits a sharp increase followed by an immediate return to baseline levels.
In contrast, change points are characterized by sustained elevations in $Z^{(t)}$ across multiple consecutive time steps, reflecting a persistent shift in the underlying network structure.
Formally, we classify a time step $t$ as a change point if both $Z^{(t)} > \tau$ and the moving average of $Z^{(t)}$ over the next $w$ time steps exceeds $\tau/2$, where $\tau$ is our anomaly threshold and $w$ is a user-defined window size.
Time steps that exceed the threshold $\tau$ but do not meet the persistence criterion are classified as event points.

\subsection{Performance Considerations}
\label{section:performance}

The computational complexity of HLSAD is dominated by the truncated singular value decomposition.
For a matrix $\mat{A} \in \mathbb{R}^{n \times n}$, computing a rank-$\ell$ decomposition requires $\mathcal{O}(n^2 \ell)$ operations.
This complexity can be significantly reduced to $\mathcal{O}(n^2 \log \ell)$ using a randomized SVD algorithm~\citep{Halko:2011}, which provides efficient approximations while maintaining high accuracy.
In practical applications, we can leverage the inherent sparsity structure of the Hodge Laplacians to achieve substantial computational efficiency gains.

Let $n_k$ denote the maximum number of $k$-simplices across all temporal instances of the simplicial complex sequence $\{\mathcal{X}^{(t)}\}_{t=1}^T$.
The computational complexity of HLSAD is then given by
\begin{align}
    \mathcal{O}\left( T \cdot \sum_{k=0}^{K} n_k^2 \log \ell \right),
\end{align}
where the summation over $k$ accounts for the spectral computations required for each Hodge Laplacian up to order $K$.

\section{Experiments}%
\label{section:experiments}

In this section, we present a comprehensive evaluation of HLSAD utilizing both synthetic and real-world datasets.
Through these experiments, we demonstrate the effectiveness of our approach in detecting both event and change points both for data-informed higher-order simplices and graph liftings.
Statistics for all datasets can be found in \cref{table:dataset-statistics} in the appendix.

To quantitatively assess the performance of our anomaly detection method, we employ the Hits@$N$ metric, which measures the proportion of correctly identified anomalies among the top $N$ most anomalous points detected by our algorithm.
Specifically, a Hits@$7$ score of $0.5$ indicates that $50\%$ of the seven points identified as most anomalous correspond to actual anomalies according to the ground truth labels.
If $N$ equals the number of ground truth anomalies, then the Hits@N score is equal to the recall.
For synthetic experiments, we utilize the ground truth labels inherent in the generation process for evaluation purposes.
In real-world datasets, we leverage known significant external events as ground truth, while acknowledging the limitation that additional unidentified external factors may have influenced the temporal evolution of the data.

\subsection{Synthetic Experiments}
\label{section:synthetic-experiments}

\begin{figure*}[t]
    \resizebox{\linewidth}{!}{%
        \begin{tikzpicture}
    \begin{groupplot}[
            width=0.21 \linewidth,
            height=105pt,
            group style={columns=4, horizontal sep=1.25cm},
            grid=major,
            grid style={dashed,gray!40},
            % xlabel=Number of Singular Values,
            xmin=0,
            xmax=600,
            xtick distance=250,
            ymin=0,
            ymax=1.05,
            ytick distance=0.2,
            no markers,
            cycle list={
                    {rwth-orange},
                    {rwth-magenta},
                    {rwth-blue}
                },
            legend image code/.code={
                    \draw[mark repeat=2,mark phase=2]
                    plot coordinates {
                            (0cm,0cm)
                            (0.15cm,0cm)        % default is (0.3cm,0cm)
                            (0.3cm,0cm)         % default is (0.6cm,0cm)
                        };%
                }
        ]

        \nextgroupplot[
            title=Hybrid,
            legend to name=synthetic-accuracy-comparison-legend,
            ylabel={Hits@$7$},
            legend columns=-1,
        ]

        % https://tex.stackexchange.com/a/2332
        \addlegendimage{empty legend}
        \addlegendentry{\scriptsize\textbf{Max Rank:}\;}

        \pgfplotsinvokeforeach{0, 1, 2} {
            \addplot+ [densely dotted, forget plot] table [
                    x=num_singular_values,
                    y=score,
                    col sep=comma,
                    discard if not={max_rank}{#1},
                    discard if not={metric}{Hits@9},
                ] {figures/synthetic-scores-graph_hybrid.csv};
            \addplot table [
                    x=num_singular_values,
                    y=score,
                    col sep=comma,
                    discard if not={max_rank}{#1},
                    discard if not={metric}{Hits@7},
                ] {figures/synthetic-scores-graph_hybrid.csv};
            \addlegendentry{#1}
        }

        \nextgroupplot[
            title=Resampled,
            ylabel={Hits@$7$},
            % move legend to the center
            xlabel=Number of Singular Values,
            every axis x label/.append style={at=(ticklabel cs:1.2)}
        ]
        \pgfplotsinvokeforeach{0, 1, 2} {
            \addplot+ [densely dotted, forget plot] table [
                    x=num_singular_values,
                    y=score,
                    col sep=comma,
                    discard if not={max_rank}{#1},
                    discard if not={metric}{Hits@9},
                ] {figures/synthetic-scores-graph_resampled.csv};
            \addplot table [
                    x=num_singular_values,
                    y=score,
                    col sep=comma,
                    discard if not={max_rank}{#1},
                    discard if not={metric}{Hits@7},
                ] {figures/synthetic-scores-graph_resampled.csv};
        }

        \nextgroupplot[title=Large, ylabel={Hits@$5$}]
        \pgfplotsinvokeforeach{0, 1, 2} {
            \addplot+ [densely dotted, forget plot] table [
                    x=num_singular_values,
                    y=score,
                    col sep=comma,
                    discard if not={max_rank}{#1},
                    discard if not={metric}{Hits@7},
                ] {figures/synthetic-scores-graph_large.csv};
            \addplot table [
                    x=num_singular_values,
                    y=score,
                    col sep=comma,
                    discard if not={max_rank}{#1},
                    discard if not={metric}{Hits@5},
                ] {figures/synthetic-scores-graph_large.csv};
        }

        \nextgroupplot[title=Triangle Closing, ylabel={Hits@$5$}]
        \pgfplotsinvokeforeach{0, 1, 2} {
            \addplot+ [densely dotted, forget plot] table [
                    x=num_singular_values,
                    y=score,
                    col sep=comma,
                    discard if not={max_rank}{#1},
                    discard if not={metric}{Hits@7},
                ] {figures/synthetic-scores-sc_triangle_closing.csv};
            \addplot table [
                    x=num_singular_values,
                    y=score,
                    col sep=comma,
                    discard if not={max_rank}{#1},
                    discard if not={metric}{Hits@5},
                ] {figures/synthetic-scores-sc_triangle_closing.csv};
        }
    \end{groupplot}

    % place a reference where the legend should appear
    % \node[anchor=north west] at ($(group c3r1.north east) + (0.7cm,0)$) {\ref{synthetic-accuracy-comparison-legend}};
    % \node[anchor=north] at ($(group c1r1.south west)!0.5!(group c4r1.south east) - (0,0.5cm)$) {\ref{synthetic-accuracy-comparison-legend}};
    \node[anchor=north east] at ($(group c4r1.south east) + (0.1cm, -0.35cm)$) {\ref{synthetic-accuracy-comparison-legend}};
\end{tikzpicture}
    }
    \caption{%
        \textbf{Performance comparison on synthetic data.}
        We compare the Hits@$N$ scores as a function of the total number of singular values used for anomaly detection.
        The solid line gives the Hits@$N$ score for $N$ the number of true anomalies in the data generation process.
        The dashed line gives the Hits@$N+2$ score, i.e., the accuracy with up to $2$ false positives.
    }
    \label{figure:synthetic-accuracy-comparison}
    \Description{%
        Four line plots showing the Hits@$N$ scores for the hybrid, resampled, large, and triangle closing synthetic experiments.
        Each plot shows the score as a function of the total number of singular values considered for anomaly detection.
    }
\end{figure*}
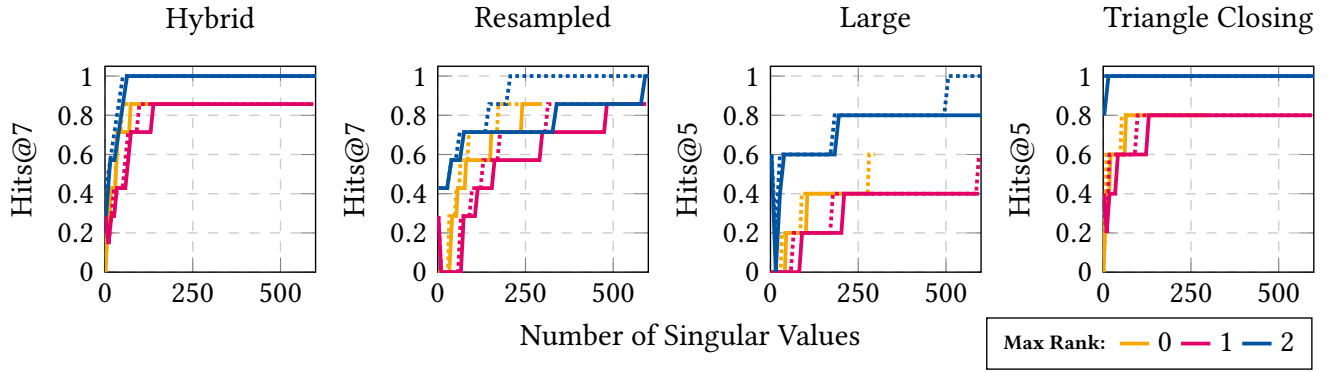

To evaluate HLSAD's effectiveness and demonstrate its advantages over graph-based approaches, we conduct four synthetic experiments:
In the first two, we replicate the \emph{hybrid} and \emph{resampled} synthetic experiments by \citet{Huang:2020}.
The time series originates from graphs sampled from a stochastic block model (SBM), which are lifted to simplicial complexes using a clique lifting.
In a third \emph{large} experiment, we evaluate scalability by using the same setting with $\num{10000}$ nodes.
In a fourth experiment, we introduce a \emph{triangle closing} dataset where, instead of using clique lifting, triangles are closed with a given probability.
This setting reflects scenarios where external events have a distinct influence on multi-way interactions in the network, which may differ from their impact on dyadic relationships.
In all settings, we change the model parameters at pre-selected time steps to simulate events and change points in the data.
The exact sampling parameters for all experiments are given in \cref{appendix:experiments}.
We use a short term and a long term window of $w_s = 5$ and $w_\ell = 10$ snapshots, respectively.

The Hits@$N$ scores for each dataset as a function on the total number of singular values computed are shown in \cref{figure:synthetic-accuracy-comparison}.
For the higher-order cases, the plotted number of singular values refers to the total number of singular values computed over all Hodge Laplacians, i.e., for a value of $300$ on the x-axis and maximum rank $K=3$, we computed $100$ singular values for each Hodge Laplacian.
In other words, all points on the same x-axis correspond to the same total number of singular values, though in the higher-order cases, these are distributed over multiple Hodge Laplacians.

The experimental results demonstrate the superior performance of HLSAD in leveraging higher-order structural information compared to the graph-based LAD baseline across both the \emph{hybrid}, \emph{large}, and \emph{triangle closing} experiments.
HLSAD consistently achieves higher Hits@$N$ scores while requiring significantly fewer singular values to attain comparable detection accuracy, indicating enhanced computational efficiency.
Specifically, in the \emph{hybrid} setting, HLSAD successfully identifies all anomalies using only $40$ singular values, whereas LAD requires more than $300$ singular values to achieve similar performance.
The advantage of HLSAD is particularly evident in the \emph{triangle closing} experiment, where it detects all anomalies with merely $10$ singular values, while LAD fails to identify anomalies that manifest exclusively in higher-order interactions due to its inherent limitation to dyadic relationships.

The interpretation of the \emph{resampled} experiment yields more nuanced results:
While the general trends observed in other experiments persist, namely superior overall performance and enhanced efficiency with fewer singular values, there exist specific instances where LAD marginally outperforms HLSAD.
Specifically, LAD achieves a Hits@$7$ score of $0.8$ with $250$ singular values, whereas HLSAD requires approximately $320$ singular values to attain equivalent performance.
However, examination of the Hits@$9$ metric reveals that HLSAD consistently and significantly outperforms LAD across all singular value quantities in a minor relaxed metric.
This suggests that while HLSAD may occasionally identify false positives, it demonstrates superior capability in detecting the complete set of true anomalies within the data.

\subsection{UCI Online Message Dataset}

\begin{table}[tb]
    \caption{%
        \textbf{Prediction scores on the UCI and Senate real-world datasets.}
        We compare HLSAD against several other state-of-the-art anomaly detection techniques.
    }
    \label{table:real-world-scores}
    \begin{tabular}{lrr}
        \toprule
        Dataset         & UCI            & Senate         \\
        Metric          & Hits@$10$      & Hits@$2$       \\
        \midrule
        HLSAD (ours)    & $\mathbf{1.0}$ & $\mathbf{1.0}$ \\
        LAD             & $0.5$          & $\mathbf{1.0}$ \\
        EdgeMonitoring  & $0.0$          & $\mathbf{1.0}$ \\
        Activity Vector & $0.5$          & $0.5$          \\
        TENSORPLAT      & $0.0$          & $0.0$          \\
        \bottomrule
    \end{tabular}
\end{table}

We further replicate the experiment on the UCI Online Message dataset~\citep{Panzarasa:2009}, which captures communication patterns within an online student community at the University of California, Irvine.
The dataset comprises a weighted network where nodes represent users and edges represent message exchanges between them.
Edge weights correspond to message lengths, quantifying interaction intensity.
The dataset encompasses $\num{1899}$ users and records $\num{59835}$ messages exchanged between April and October 2004.
On June 19, the spring term ended and on September 20, the fall term began, marking significant external events that influenced the communication patterns.
We construct daily network snapshots for our temporal analysis and use a clique lifting to transform the dyadic interactions into higher-order simplices.
Following the arguments outlined in the previous works, we employ a short-term context window of $7$ days, with a corresponding long-term window of $14$ days to capture broader temporal patterns.

Our analysis, summarized in \cref{table:real-world-scores}, demonstrate the superior performance of HLSAD in detecting temporal anomalies.
The method achieves a perfect Hits@$10$ score, accurately identifying all ground truth anomalies in the dataset.
LAD correctly detected the end of the spring term, but missed the beginning of the fall term by a single timestep.
Similarly, the Activity Vector method only successfully identified the end of the spring term.
Both EdgeMonitoring and TENSORPLAT prove ineffective, failing to identify any anomalies in the dataset.
Notably, all methods that demonstrate meaningful detection capabilities---HLSAD, LAD, and Activity Vector---utilize spectral-based approaches, underlining the effectiveness of spectral features in temporal anomaly detection tasks.

\subsection{Senate Network}

We analyze the Senate co-sponsorship network~\citep{Fowler:2006,Fowler:2006b}, which represents legislative collaboration patterns through higher-order interactions: when multiple senators co-sponsor a bill, they form a $k$-simplex in the complex. The dataset comprises biannual snapshots spanning multiple congressional sessions.

Prior research has identified two notable anomalies in collaboration patterns: the 100th Congress exhibited exceptionally high levels of bipartisan cooperation, while the 104th Congress marked a historical low point in cross-party collaboration~\citep{Wang:2017}.
Our experimental results demonstrate that HLSAD successfully identifies both these significant structural anomalies using only the top $6$ singular values of the Hodge Laplacians, as shown in \cref{table:real-world-scores}.

\subsection{MIT Reality Mining}

\begin{figure}[tb]
    \resizebox{\linewidth}{!}{%
        \begin{tikzpicture}
    \begin{groupplot}[
            width=0.5 \linewidth,
            height=100pt,
            group style={group size=2 by 1, horizontal sep=1.5cm},
            grid=major,
            grid style={dashed,gray!40},
            xlabel=Delay,
            xmin=0,
            xmax=5,
            ymin=0.5,
            ymax=1.05,
            ytick distance=0.2,
            no markers,
            legend style={legend columns=-1, column sep=7pt}
        ]
        \nextgroupplot[ylabel=Precision, legend to name=mit-grouplegend]
        \addplot [index of colormap=2] coordinates {(0,0.75) (1,0.8125) (2,0.99) (3,0.99) (4,0.99) (5,0.99)};
        \addlegendentry{LetoChange}
        \addplot [index of colormap=1] coordinates {(0,0.75) (1,0.8125) (2,0.9375) (3,0.9375) (4,0.985) (5,0.985)};
        \addlegendentry{LAD}
        \addplot [index of colormap=0] coordinates {(0,0.75) (1,0.875) (2,1.0) (3,1.0) (4,1.0) (5,1.0)};
        \addlegendentry{HLSAD}

        \nextgroupplot[ylabel=Recall]
        \addplot [index of colormap=2] coordinates {(0,0.6667) (1,0.8889) (2,0.9444) (3,0.9444) (4,0.9444) (5,1.0)};
        \addplot [index of colormap=1] coordinates {(0,0.65) (1,0.943) (2,0.99) (3,0.99) (4,0.99) (5,0.99)};
        \addplot [index of colormap=0] coordinates {(0,0.6667) (1,0.9444) (2,1.0) (3,1.0) (4,1.0) (5,1.0)};
    \end{groupplot}
    \node[anchor=north] at ($(group c1r1.south west)!0.5!(group c2r1.south east) - (0.5cm, 0.9cm)$) {\ref{mit-grouplegend}};
\end{tikzpicture}
    }
    \caption{%
        \textbf{Precision and recall as a function of the detection delay on the MIT Reality Mining dataset.}
        We compare our method HLSAD against the spectral-based LAD approach and the model-based LetoChange framework.
    }
    \label{figure:mit-results}
    \Description{%
        Two line plots showing the precision and recall of HLSAD, LAD, and LetoChange on the MIT Reality Mining dataset.
        The precision and recall are shown as a function of the detection delay, which is the time difference between when the ground truth event happened and when the algorithm has detected the anomaly.
    }
\end{figure}
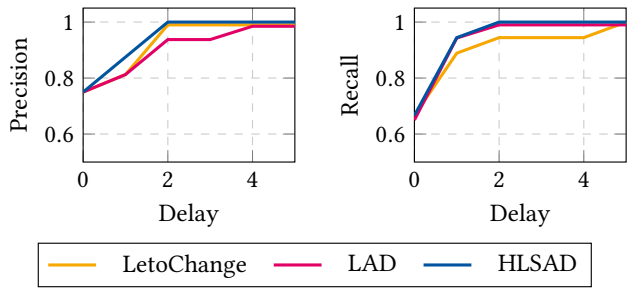

For our third real-world experiment, we analyze the MIT Reality Mining dataset~\citep{Eagle:2006}, which documents proximal interactions between students through continuous Bluetooth device monitoring.
We construct weekly temporal networks from the raw proximity data, where edges indicate detected device co-location between participants.
Higher-order simplices are derived though clique lifting.
To ensure data quality, we exclude participants with incomplete or corrupted records, particularly when data was corrupted due to power failures.
For ground truth validation, we leverage the $16$ external events identified by \citet{Peel:2015} that potentially influenced interaction patterns throughout the study period.

In analogy to previous work~\citep{Peel:2015}, we compute the precision and recall of our method as a function of the detection delay between the ground truth events $\{ t_j \}$ and the detected anomaly $\{ \hat{t}_i \}$.
More formally, the precision and recall are defined as
\begin{align}
    \operatorname{Precision}(s) & = \frac{1}{n_\text{pred}} \sum_{i} \delta \left( \inf_j (\hat{t}_i - t_j) \leq s \right)  \\
    \operatorname{Recall}(s)    & = \frac{1}{n_\text{true}} \sum_{j} \delta \left( \inf_i (\hat{t}_i - t_j) \leq s \right),
\end{align}
where $s$ is the allowed detection delay, $n_\text{pred}$ is the number of predicted anomalies, $n_\text{true}$ is the number of true anomalies, and $\delta(\cdot)$ is the indicator function.

In our experimental evaluation, we compare HLSAD against two established baselines: the spectral-based LAD approach~\citep{Huang:2020} and the model-based LetoChange framework~\citep{Peel:2015}.
The comparative analysis, presented in \cref{figure:mit-results}, demonstrates that HLSAD achieves comparable precision to LetoChange for instantaneous anomaly detection ($s=0$), while exhibiting superior detection capabilities when allowing for some detection delays.
Furthermore, HLSAD consistently outperforms LetoChange in terms of recall across all detection delays, indicating superior sensitivity in identifying true anomalies within the dataset.
Although the spectral-based LAD method exhibits recall performance comparable to HLSAD, it generates additional false positives in temporal proximity to ground truth events, resulting in diminished precision metrics.
Overall, our results highlight the effectiveness of HLSAD in detecting temporal anomalies in real-world social interaction networks.

\section{Conclusion}%
\label{section:conclusion}

In this paper, we introduced a novel methodology for detecting anomalies in time-evolving simplicial complexes.
Our approach leverages the spectral properties of Hodge Laplacians to capture higher-order structural changes that traditional graph-based methods fail to detect.
Through comprehensive experiments on both synthetic and real-world datasets, we demonstrated that our method successfully identifies both event and change points in temporal sequences of simplicial complexes.

The key advantages of HLSAD are threefold:
First, it naturally handles higher-order interactions that are either inherent in the data or arise through graph lifting techniques.
Second, it maintains computational efficiency through the use of truncated SVD computations on sparse Hodge Laplacians.
Third, it is invariant to node permutations and growing simplicial complexes, making it applicable when consistent node ordering cannot be guaranteed or a priori knowledge of the total number of nodes is unavailable.

Our experimental results show that our method achieves superior detection accuracy compared to traditional graph-based approaches, particularly when analyzing datasets with significant higher-order structure.
Furthermore, when applied to lifted graph data, our method requires fewer eigenvalues to achieve comparable accuracy, resulting in improved computational efficiency.

Future work could explore adaptive window sizes for the temporal context, automated parameter selection techniques, and extensions to weighted or directed simplicial complexes.
Additionally, investigating the relationship between specific types of structural changes and their manifestation in the Hodge spectrum could provide deeper insights into the nature of detected anomalies as well as possibilities to fine-tune to specific anomaly kinds.

In many real-world applications, the presence of attributed data or node features can provide additional context for anomaly detection beyond the structural information captured by the Hodge Laplacian.
However, our current implementation focuses solely on the topological structure of the simplicial complex.
In future work, we plan to extend our method to incorporate attributed data or node features, allowing for a more comprehensive analysis of anomalies in complex systems.

\begin{acks}
    The authors acknowledge funding by the \grantsponsor{nrw}{Ministry of Culture and Science of the German State of North Rhine-Westphalia}\, (“NRW Rückkehrprogramm”) and the European Union (\grantsponsor{high-hopes}{ERC}, HIGH-HOPeS, \grantnum{high-hopes}{101039827}).
    Views and opinions expressed are however those of the authors only and do not necessarily reflect those of the European Union or the European Research Council Executive Agency.
    Neither the European Union nor the granting authority can be held responsible for them.
\end{acks}

\bibliographystyle{ACM-Reference-Format}
\bibliography{references}

%%%%%%%%%%%%%%%%%%%%%%%%%%%%%%%%%%%%%%%%%%%%%%%%%%%%%%%%%%%%%%%%%%%%%%%%%%%%%%%
%%%%%%%%%%%%%%%%%%%%%%%%%%%%%%%%%%%%%%%%%%%%%%%%%%%%%%%%%%%%%%%%%%%%%%%%%%%%%%%
% APPENDIX
%%%%%%%%%%%%%%%%%%%%%%%%%%%%%%%%%%%%%%%%%%%%%%%%%%%%%%%%%%%%%%%%%%%%%%%%%%%%%%%
%%%%%%%%%%%%%%%%%%%%%%%%%%%%%%%%%%%%%%%%%%%%%%%%%%%%%%%%%%%%%%%%%%%%%%%%%%%%%%%

\appendix

\section{Teaser Figure}

\Cref{table:teaser-parameters} lists the exact parameters for the time series illustrated in \cref{figure:teaser}.
Graph skeletons are sampled with $30$ nodes from a SBM using uniform community sizes.
Triangles are closed (identified with a $2$-simplex) with probability $p_\Delta$.
The continuity rate $\alpha^{(t)}$ is set to $1.0$ for anomaly points and $\alpha^{(t)} = 0.005$ everywhere else.

\begin{table}[H]
    \caption{%
        SBM parameters and triangle closing probability for the time series used in \cref{figure:teaser}.
        Parameter changes are marked in bold.
    }
    \label{table:teaser-parameters}
    \begin{tabular}{rlrrrr}
        \toprule
        Step   & Type   & Communities & Intra           & Inter           & Triangle        \\
               &        & $N_c$       & $p_\text{in}$   & $p_\text{ex}$   & $p_\Delta$      \\
        \midrule
        1--9   & -      & $3$         & $0.25$          & $0.05$          & $0.09$          \\
        10     & Event  & $3$         & $\mathbf{0.15}$ & $0.05$          & $0.09$          \\
        11--19 & -      & $3$         & $0.25$          & $0.05$          & $0.09$          \\
        20-29  & Change & $3$         & $0.25$          & $\mathbf{0.15}$ & $0.09$          \\
        30-39  & Change & $3$         & $0.25$          & $0.15$          & $\mathbf{0.03}$ \\
        \bottomrule
    \end{tabular}
\end{table}

\section{Experimental Setup}%
\label{appendix:experiments}

\Cref{table:synthetic-hybrid-resampled-parameters} lists the exact SBM parameters for the \emph{hybrid} and \emph{resampled} synthetic experiments, which are derived from the work of \citet{Huang:2020}.
We excluded their \emph{pure} experiment, as anomalous points in this setting can be perfectly detected in the first singular values.
\Cref{table:synthetic-large-parameters} and \ref{table:synthetic-experiments-triangle-parameters} lists the parameters for the \emph{large} and \emph{triangle closing} synthetic experiment, respectively.

All graphs except for the large dataset are sampled with $500$ nodes and uniform community sizes.
For the large dataset, the sampled graphs have $\num{10000}$ nodes.
For the hybrid and large setting, the continuity rate was set to $\alpha^{(t)} = 1.0$ for anomaly points and $\alpha^{(t)} = 0.1$ everywhere else.
For the resampled setting, the continuity rate was set to $\alpha^{(t)} = 1.0$ everywhere, i.e., a fresh graph was resampled at each time step.
For the first three experiments, we used a clique lifting to obtain the simplicial complexes from the graph skeletons sampled from the SBMs.
For the \emph{triangle closing} experiment, triangles are closed with a given probability $p_\Delta$, simulating a data-informed higher-order structure where external events have a distinct influence on multiway interactions in the network.

\begin{table}[H]
    \caption{%
        \textbf{SBM parameters for the \emph{hybrid} and \emph{resampled} synthetic experiments.}
        Parameter changes are marked in bold.
    }
    \label{table:synthetic-hybrid-resampled-parameters}
    \begin{tabular}{rlrrr}
        \toprule
        Step     & Type   & Communities   & Intra-Prob.     & Inter-Prob.     \\
                 &        & $N_c$         & $p_\text{in}$   & $p_\text{ex}$   \\
        \midrule
        1--16    & -      & $4$           & $0.25$          & $0.05$          \\
        17       & Event  & $4$           & $0.25$          & $\mathbf{0.15}$ \\
        18--31   & -      & $4$           & $0.25$          & $0.05$          \\
        32--61   & Change & $\mathbf{10}$ & $0.25$          & $0.05$          \\
        62       & Event  & $10$          & $0.25$          & $\mathbf{0.15}$ \\
        63--76   & -      & $10$          & $0.25$          & $0.05$          \\
        77--91   & Change & $\mathbf{2}$  & $\mathbf{0.5}$  & $0.05$          \\
        92       & Event  & $2$           & $0.5$           & $\mathbf{0.15}$ \\
        93--106  & -      & $2$           & $0.5$           & $0.05$          \\
        107--136 & Change & $\mathbf{4}$  & $\mathbf{0.25}$ & $0.05$          \\
        137      & Event  & $4$           & $0.25$          & $\mathbf{0.15}$ \\
        138--150 & -      & $4$           & $0.25$          & $0.05$          \\
        \bottomrule
    \end{tabular}
\end{table}

\begin{table}[H]
    \caption{%
        \textbf{SBM parameters for the \emph{large} synthetic experiment.}
        Parameter changes are marked in bold.
    }
    \label{table:synthetic-large-parameters}
    \begin{tabular}{rlrrr}
        \toprule
        Step    & Type   & Communities   & Intra-Prob.      & Inter-Prob.        \\
                &        & $N_c$         & $p_\text{in}$    & $p_\text{ex}$      \\
        \midrule
        1--16   & -      & $4$           & $0.0125$         & $0.0025$           \\
        17      & Event  & $4$           & $0.0125$         & $\mathbf{0.0075}$  \\
        18--31  & -      & $4$           & $0.0125$         & $0.0025$           \\
        32--61  & Change & $\mathbf{10}$ & $0.25$           & $0.0025$           \\
        62      & Event  & $10$          & $0.25$           & $\mathbf{0.0075}$  \\
        63--76  & -      & $10$          & $0.25$           & $0.0025$           \\
        77--91  & Change & $\mathbf{2}$  & $\mathbf{0.025}$ & $0.0025$           \\
        92      & Event  & $2$           & $0.025$          & $\mathbf{0.00755}$ \\
        93--100 & -      & $2$           & $0.025$          & $0.0025$           \\
        \bottomrule
    \end{tabular}
\end{table}

\begin{table*}[t]
    \caption{%
        \textbf{Statistics for the datasets used in our experiments.}
        We give the mean and standard deviation of the number of nodes, edges, and triangles in the dynamic simplicial complexes.
    }
    \label{table:dataset-statistics}
    \begin{tabular}{lSSSS}
        \toprule
        Dataset          & {$T$} & {$\avg | \mathcal{X}_0^{(t)} |$} & {$\avg | \mathcal{X}_1^{(t)} |$} & {$\avg | \mathcal{X}_2^{(t)} |$} \\
        \midrule
        Hybrid           & 150   & 500 +- 0                         & 19266 +- 10346                   & 190476 +- 284845                 \\
        Resampled        & 150   & 500 +- 0                         & 19237 +- 10254                   & 188018 +- 279797                 \\
        Triangle Closing & 60    & 500 +- 0                         & 16898 +- 6652                    & 22092 +- 31509                   \\
        Large            & 100   & 10000 +- 0                       & 394171 +- 215818                 & 211516 +- 294775                 \\
        \midrule
        UCI              & 193   & 117 +- 121                       & 134 +- 180                       & 5 +- 11                          \\
        Senate           & 12    & 101 +- 0.8                       & 5033 +- 88                       & 165770 +- 4366                   \\
        Reality Mining   & 32    & 94 +- 17                         & 603 +- 302                       & 2937 +- 2214                     \\
        \bottomrule
    \end{tabular}
\end{table*}

\begin{table}[!ht]
    \caption{%
        \textbf{SBM parameters and triangle closing probability for the \emph{triangle closing} synthetic experiment.}
        Changes in the parameters are marked in bold.
    }
    \label{table:synthetic-experiments-triangle-parameters}
    \begin{tabular}{rlrrrr}
        \toprule
        Step   & Type   & Communities & Intra          & Inter           & Triangle       \\
               &        & $N_c$       & $p_\text{in}$  & $p_\text{ex}$   & $p_\Delta$     \\
        \midrule
        1--9   & -      & $4$         & $0.25$         & $0.05$          & $0.8$          \\
        10     & Event  & $4$         & $0.25$         & $\mathbf{0.15}$ & $0.8$          \\
        11--19 & -      & $4$         & $0.25$         & $0.05$          & $0.8$          \\
        20--29 & Change & $4$         & $0.25$         & $0.05$          & $\mathbf{0.7}$ \\
        30     & Event  & $4$         & $0.25$         & $\mathbf{0.15}$ & $0.7$          \\
        31--39 & -      & $4$         & $0.25$         & $0.05$          & $0.7$          \\
        40--49 & Change & $4$         & $\mathbf{0.5}$ & $0.05$          & $\mathbf{0.5}$ \\
        50--59 & Change & $4$         & $0.5$          & $\mathbf{0.05}$ & $0.5$          \\
        \bottomrule
    \end{tabular}
\end{table}

\Cref{table:dataset-statistics} lists the statistics for the synthetic and real-world datasets used in our experiments.
We implemented HLSAD in Python, leveraging the TopoX library~\citep{Hajij:2024} for topological computations and SciPy~\citep{Virtanen:2020} for efficient algebraic operations.
HLSAD has been implemented as a Snakemake pipeline~\citep{Mölder:2021}.
We took inspiration from the implementation of LAD by \citet{Huang:2020}.
For comparative analysis, we utilized reference implementations of baseline methods, except LAD, which we realized as a special case of HLSAD with appropriate parameter settings.
All source code and experimental data will be made publicly available with the camera-ready version to ensure reproducibility.

\section{Ablation Study: Influence of the Context Window}
\label{appendix:context-window}

\begin{figure}[h!]
    \resizebox{\linewidth}{!}{%
        \begin{tikzpicture}
    \begin{axis}[
            width=0.65 \linewidth,
            xlabel=Short Window Size $w_s$,
            ylabel=Hits@$N$,
            legend pos=outer north east,
            legend cell align=left,
            cycle list={
                    {rwth-magenta, mark=*},
                    {rwth-blue, mark=square*},
                    {rwth-orange, mark=diamond*},
                    {rwth-violett, mark=triangle*},
                },
        ]
        % https://tex.stackexchange.com/a/2332
        \addlegendimage{empty legend}
        \addlegendentry{\textbf{Dataset}}

        \addplot table [
                x=short_window,
                y=score,
                col sep=comma,
                discard if not={metric}{Hits@7},
            ] {figures/synthetic-ablation-study-graph_hybrid.csv};
        \addlegendentry{Hybrid}

        \addplot table [
                x=short_window,
                y=score,
                col sep=comma,
                discard if not={metric}{Hits@7},
            ] {figures/synthetic-ablation-study-graph_resampled.csv};
        \addlegendentry{Resampled}

        \addplot table [
                x=short_window,
                y=score,
                col sep=comma,
                discard if not={metric}{Hits@5},
            ] {figures/synthetic-ablation-study-sc_large.csv};
        \addlegendentry{Large}

        \addplot table [
                x=short_window,
                y=score,
                col sep=comma,
                discard if not={metric}{Hits@5},
            ] {figures/synthetic-ablation-study-sc_triangle_closing.csv};
        \addlegendentry{Triangle Closing}
    \end{axis}
\end{tikzpicture}
    }
    \caption{%
        \textbf{Ablation study on the influence of the context window on the performance of HLSAD.}
        We show the Hits@$N$ scores for different context window sizes on the synthetic datasets.
        The long window $w_\ell$ is set to double the short window $w_s$.
    }
    \label{figure:ablation-context-window}
    \Description{%
        Line plot showing the prediction accuracy of HLSAD on the synthetic datasets as a function of the context window size.
        The x-axis shows the context window size, the y-axis shows the Hits@$N$ score.
        For most datasets, the score drops for large context windows.
        For some datasets, the score also drops for small context windows.
    }
\end{figure}
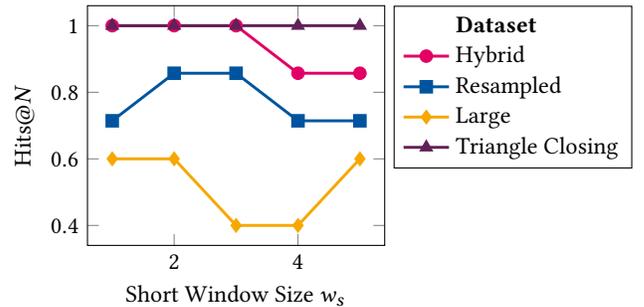

In this section, we conduct an ablation study to investigate the influence of the context window sizes on the prediction performance of HLSAD.
The context window sizes have only small influence on the computation time of HLSAD, as the computational complexity is dominated by the SVD computation, which is independent of the context window sizes, see \cref{section:performance}.

To that end, \cref{figure:ablation-context-window} shows the Hits@$N$ scores for different context window sizes on the synthetic datasets used in \cref{section:synthetic-experiments}.
In this experiment, we set the number of singular values to $75$ and the maximal simplex rank to $2$.
We can see that the Hits@$N$ scores drop for large context windows.
This is likely due to the fact that large contexts smooth out any changes in the network structure, i.e., $\vec{\sigma}^{(t)}$ and $\vec{\sigma}^{(t+1)}$ are more similar with larger context even if the network structure changes.
On the other hand, short windows can also be problematic, see the resampled setting, as the infered typical spectra tend to be noisy and lead to more false positives.
Interestingly, the \emph{large} dataset shows a counterintuitive behavior, where the accuracy drops in the middle and increases at its tails.
We don't have an intuitive explanation for this behavior.

The results show that the context window sizes can have significant influence on the performance of HLSAD, which is in line with other algorithms for temporal graph anomaly detection employing context windows.
Moreover, the different behaviors even on the relatively similar synthetic datasets highlight the importance to fine-tune the context window size specifically for the dataset at hand.

\end{document}